# DCSO: Dynamic Combination of Detector Scores for Outlier Ensembles

Yue Zhao
Department of Computer Science
University of Toronto
Toronto, Canada
yuezhao@cs.toronto.edu

Maciej K. Hryniewicki
Data Assurance & Analytics
PricewaterhouseCoopers
Toronto, Canada
maciej.k.hryniewicki@pwc.com

## ABSTRACT
Selecting and combining the outlier scores of different base detectors used within outlier ensembles can be quite challenging in the absence of ground truth. In this paper, an unsupervised outlier detector combination framework called DCSO is proposed, demonstrated and assessed for the dynamic selection of most competent base detectors, with an emphasis on data locality. The proposed DCSO framework first defines the local region of a test instance by its $k$ nearest neighbors and then identifies the top-performing base detectors within the local region. Experimental results on ten benchmark datasets demonstrate that DCSO provides consistent performance improvement over existing static combination approaches in mining outlying objects. To facilitate interpretability and reliability of the proposed method, DCSO is analyzed using both theoretical frameworks and visualization techniques, and presented alongside empirical parameter setting instructions that can be used to improve the overall performance.

## 1. INTRODUCTION
Outlier detection methods aim to identify anomalous data points from normal ones and are useful in many applications including the detection of anomalous behavior in social media [10] as well as the detection of faulty mechanical devices [9]. Over the years, numerous unsupervised outlier detection methods have been proposed [7, 21–23] because the ground truth is often absent in outlier mining [1]. In spite of some recent successes in outlier detection, unsupervised approaches have been criticized for yielding both high false positive and high false negative rates [12]. To improve the detection accuracy and stability, researchers have recently devoted their efforts to the application of ensemble methods to outlier detection problems [1, 2, 31, 38], and several new outlier ensemble algorithms have been proposed [22, 23, 30, 31, 37]. Ensemble learning uses combinations of various base estimators to achieve more reliable and superior results than those attainable with an individual estimator [14, 19]. This approach typically involves two key stages [31]: (i) the **Generation** stage, which creates a pool of base estimators and (ii) the **Combination** stage, which synthetizes the base estimators to create an improved final output. The model combination is particularly important, as there is often an inherent risk that some of the constituent estimators could potentially deteriorate the capabilities of the ensemble, rather than improve them [29, 30]. Although existing outlier ensemble methods show promising results, there is still room for improvement as the model combination stage could be quite challenging at times [1].

First of all, most existing combination methods are fully static and do not involve any detector selection processes, even though they are critical in detector combination for outlier ensembles [30]. The lack of detector selection limits the benefits of model combination as base detectors may not be fully capable of identifying all of the unknown outlier instances [11]. Although static averaging of all base detector scores is the most widely used method, it could possibly have the high-performing detectors being neutralized by the low-performing ones in overall model performance [2].

Secondly, the importance of data locality is often underestimated and rarely discussed in detector selection and combination. Specifically, the competency of a base detector is typically evaluated globally on all training data points instead of being on the local region related to the test object. For instance, a popular combination method called weighted averaging [38] uses the Pearson correlation between the detector score and the pseudo ground truth on all training points as the detector weight [38]. Numerous local detection algorithms [7, 21, 33] have therefore been developed as of late, under the premise that certain types of outliers are more easily identified by local data relationships [33]. As a result, considering the base detector performance in the local region of a test instance may be helpful to detector selection and combination.

Thirdly, limited interpretability and reliability of unsupervised combination frameworks prevent these methods from being used in mission-critical tasks. Some possible causes include: (i) the lack of ground truth impedes a controlled combination process; (ii) the combination frameworks may involve random processes leading to unstable performance and poor reproducibility; (iii) only the combination result is available, but the decision procedure is untraceable and (iv) algorithm results are often analyzed by direct comparison instead of statistical analysis.

To address the aforementioned limitations, a fully unsupervised framework called DCSO (**D**ynamic **C**ombination of Detector **S**cores for **O**utlier Ensembles) is proposed in this research to select and combine base detectors dynamically with a focus on data locality. The idea is motivated by an established supervised ensemble framework known as Dynamic Classifier Selection (DCS) [19]. DCS selects the best classifier for each test instance on the fly by evaluating base classifiers' competency on the local region of a test instance [11]. The rationale behind this is that not every base classifier is good at categorizing all unknown test instances and may be more likely to specialize in different local regions [11]. Similarly, DCSO first defines the local region of a test instance as its $k$ nearest training points, and then identifies the most competent base detector in the local region by the similarity to the pseudo ground truth. To further improve algorithm stability and capacity, ensemble variations of DCSO are proposed: multiple promising detectors are kept for a second-phase combination instead of only using the most competent detector. To the best of the authors' knowledge, this is the first published effort to adapt DCS from supervised classification tasks to unsupervised outlier ensembles.

The proposed DCSO framework has the following advantages:

1. DCSO outperforms traditional static methods on most datasets with a significant improvement in precision;

2. DCSO has great extensibility as it is compatible with different types of base detectors, such as Local Outlier Factor (LOF) [7] and $k$ Nearest Neighbors ($k$NN);

3. DCSO can show the combination process for each test instance by providing the selected base detector(s), which helps model validation and reproducibility.

To improve model interpretability and deconstruct the black-box nature of outlier detector combination, various analysis methods are employed herein. First, a theoretical explanation is provided under a recently proposed framework by Aggarwal and Sathe [3]. Second, visualization techniques are leveraged to intuitively explain why DCSO works and when best to use it. Third, statistical tests are used to reliably evaluate algorithm performance. And fourth, the effect of parameters is discussed alongside an empirical instruction setting. In summary, DCSO is easy to understand, stable to use and effective for unsupervised outlier detector combination. All source codes, experiment results and figures are openly shared for reproduction[1].

## 2. RELATED WORK
### 2.1 Dynamic Classifier Selection and Dynamic Ensemble Selection

Dynamic Classifier Selection (DCS) is a representative Multiple Classifier Systems framework. The idea was first proposed by Ho et al. in 1994 [19], and extended by Woods et al. to DCS Local Accuracy in 1997 [36] that selects the best base classifier by its accuracy in the local region. The rationale is that base classifiers may have distinct errors and some degree of complementarity [8]. Selecting and combining various base classifiers dynamically leads to performance improvement over static ensembles such as majority vote of all base classifiers. Both Ho and Wood's work illustrates DCS' superior performance in real-world applications. From a theoretical perspective, Giacinto and Roli proved that under certain assumptions, the optimal Bayes classifier could be obtained by selecting non-optimal classifiers, as the foundation of DCS [17]. The idea of DCS is further expanded by Ko, Sabourin and Britto [20] to Dynamic Ensemble Selection (DES). Compared with DCS, DES picks multiple base classifiers for each test instance for a second-phase combination. As the difficulty of identifying different test patterns varies, selecting a group of classifiers should be more stable than only selecting one of the best. DES helps distribute the risk over a group of classifiers instead of an individual classifier. Experimental results confirm that DES is more stable than DCS [20]. Motivated by the idea of DCS and DES, DCSO has been designed by adapting both algorithms to unsupervised outlier detector combination tasks.

### 2.2 Data Locality in Outlier Detection

The relationship among data objects is critical in outlier detection; anomaly mining algorithms could be roughly categorized as global versus local [21, 31, 32]. The former makes the decision using all objects, while the latter only considers a local selection of objects [32]. In both cases, their applicability is data-dependent. For instance, global outlier algorithms are useful when the outliers are far from the rest of the data [30] but may fail to find objects that are outliers in local neighborhoods, as is often the case with high-dimension datasets [7, 33]. Additionally, assuming that a data object is relative to all other objects could lead to low accuracy for the data generated from a mixture of distributions, where the global characteristic is rather irrelevant [33]. Many local algorithms have been therefore proposed, such as LOF [7], LoOP [21] and Gloss [33]. However, the importance of data locality has been rarely considered in outlier detector selection and combination. Most of the detector evaluation methods directly or indirectly depend on all training data points, e.g., the weight calculation in weighted averaging [38]. Therefore, it is reasonable to consider the data locality of the test instance while selecting and combining constituent detectors, as different base detectors may only work in certain local regions. DCSO considers both global and local data relationships: base detectors are trained with the entire dataset globally, while the detector selection and combination focus on data locality.

### 2.3 Outlier Score Combination

Recently, outlier ensemble has become a popular research area [1–3, 38] and numerous methods have been proposed, including: (i) parallel methods such as Feature bagging [22] and Isolation Forest [23]; (ii) sequential methods including CARE [31] and SELECT [30] and (iii) hybrid approaches like BORE [25] and XGBOD [37]. In classification tasks, ensemble methods can be categorized as bagging [6], boosting [15] and stacking [35]. When the ground truth exists, base detector selection and combination can be guided by the label, in which the supervised approach is also applicable [2]. Given a small number of labels are available, semi-supervised approaches are helpful for model combination, in which unsupervised methods can be used as representation extractors to improve supervised detection methods [2]; corresponding algorithms have also been proposed in [25, 37].

When the ground truth is unavailable, combining outlier models is important yet challenging, especially in bagging [3]. One of the earliest works, Feature Bagging [22], constructs diversified base detectors by training on randomly selected subsets of features, and combines the outlier scores statically. Widely used unsupervised combination algorithms in bagging are often both static and global (SG), e.g., averaging base detectors scores. A list of representative SG methods are described below (see [1–3, 31, 38] for details):

1. Averaging (*SG_A*): averaging the scores of all base detectors as the final outlier score of a test object.

2. Maximization (*SG_M*): reporting the maximum outlier score across all base detectors regarding a test object.

3. Threshold Sum (*SG_THRESH*): discarding all outlier scores below a threshold (e.g., removing all negative scores) and summing over the remaining base detector scores.

---
[1] https://github.com/yzhao062/DCSO

4. Average-of-Maximum (*SG_AOM*): after dividing base detectors into subgroups, taking the maximum score for each subgroup as the subgroup score. The final score is calculated by averaging all subgroup scores.

5. Maximum-of-Average (*SG_MOA*): after dividing base detectors into subgroups, taking the average score for each subgroup as the subgroup score. The final score is calculated as the maximum among all subgroup scores.

6. Weighted Averaging (*SG_WA*): generating the pseudo training ground truth by averaging all base detector scores. The weight of each base detector is calculated as the Pearson correlation between its training score and the pseudo ground truth. Pearson Correlation in Eq. (1) measures the similarity between two vectors ***p*** and ***q*** (*l* denotes vector length), where $\bar{p}$ and $\bar{q}$ are the mean of the vectors. Once the detector weights are generated, the final score is the weighted average of all detector scores.

$$\rho(\boldsymbol{p},\boldsymbol{q}) = \frac{\sum_{i=1}^{l}(p_i - \bar{p})(q_i - \bar{q})}{\sqrt{\sum_{i=1}^{l}(p_i - \bar{p})^2}\sqrt{\sum_{i=1}^{l}(q_i - \bar{q})^2}} \quad (1)$$

As discussed in Section 2.2, SG methods ignore the importance of data locality while evaluating and combining detectors, which may be inappropriate given the characteristics of outliers [7, 33]. Moreover, the absence of detector selection may have inaccurate detectors retained, causing an adverse effect [30, 31]. Taking *SG_A* as an example, scores are averaged with equal weights. Inevitably, high-performing detectors are negated by low-performing ones [2]. *SG_M* is rather heuristic to result in unstable results [3]. For *SG_AOM* and *SG_MOA*, they have a second-phase combination to improve the model capacity and stability. However, it is not easy to understand and interpret the model when randomness is inherently part of the procedure and limited traceability of the base detectors' contribution is a result.

Selective detector combination can be beneficial for unsupervised outlier ensembles [30] by addressing the limitations of SG methods. There have been several attempts to build outlier ensembles dynamically and sequentially in a boosting style. Rayana and Akoglu introduced SELECT [30] and CARE [31] to pick promising detectors and eliminate the underperforming ones. SELECT generates the pseudo ground by averaging detectors' outlier probabilities; the weighted Pearson correlation between the pseudo ground truth and the base detector outlier probability is then used to decide whether to keep the detector. SELECT shows great potential for both temporal graphs and multi-dimensional outlier data. In this study, DCSO is designed to fill the gaps of SG methods by stressing the importance of data locality and dynamic detector selection; all aforementioned SG algorithms are thus included as baselines. It should be noted that the purpose of DCSO is not to outperform the best base outlier detector when the ground truth is missing, but rather to explore the use of dynamic combination in outlier ensembles for the sake of improved accuracy, stability and interpretability.

## 2.4 Model Interpretability and Reliability

Outlier detection is critical in many real-world applications. However, users often express the concerns regarding outlier models interpretability [21]. To better analyze and demonstrate the mechanism of outlier detection methods, researchers have used both theoretical and practical explanations. Recently, Aggarwal and Sathe laid the theoretical foundation for outlier ensemble [3] using Bias-variance tradeoff, a widely used framework for analyzing the generalization error in classification problems [37]. It is evident that outlier detection could be viewed as a special case of binary classification with skewed classes, where the minority class represents outliers [30, 31]. Similar to classification problems, the reducible generalization error may be minimized by either reducing squared bias or variance in outlier ensemble, where the tradeoff appears. A low bias detector is sensitive to the data variation with high instability; a low variance detector is less sensitive to data variation but may fit complex data poorly. The goal of outlier ensemble is to control both bias and variance to reduce the overall generalization error. Various newly proposed algorithms have been analyzed using this new framework to enhance interpretability and reliability [30, 31, 37].

Practical approaches, such as visualizations and interactive applications make it easier for users to understand the model. Both Micenková, McWilliams and Assent [25] as well as Zhao and Hryniewicki [37] used t-SNE visualization [24] to show their methods succeed in separating outliers from the normal points. In addition, Perozzi and Akoglu [28] proposed an interactive visual exploration and summarization tool to provide interpretable results to users for identifying communities and anomalies in attributed graphs. Das et al. [12] introduced an interactive method to incorporating expert feedback into the anomaly detection. These visual and interactive methods help users understand and trust the model in a more intuitive way. Moreover, data mining algorithms have rarely been analyzed through quantitative methods such as statistical tests [13]. In this study, both theoretical and visualization techniques are used to further improve the DCSO framework's inherently high interpretability, and statistical tests are carried out to assess the model performance reliably.

## 3. ALGORITHM DESIGN

As classification ensembles, DCSO has two key stages. In the ***Generation*** stage, the chosen base detector algorithm is initialized with distinct parameters to build a pool of diversified detectors, and all are then fitted on the entire training dataset. In the ***Combination*** stage, DCSO picks the most competent detector in the local region defined by the test instance. Finally, the selected detector is used to predict the outlier score for the test instance.

### 3.1 Base Detector Generation

An effective ensemble should be constructed with diversified base estimators [31, 38]; the diversity among base estimators helps different data characteristics to be learned. If base detectors are highly correlated, the benefit of model combination is limited. With a group of homogeneous base detectors, the diversity can be induced by training on different subsamples, using various subsets of features or varying model parameters [8, 38]. DCSO uses distinct initial parameters to construct a pool of diversified base models when the same type of base detector is chosen.

Let $X_{train} \in \mathbb{R}^{n \times d}$ denote the training data with *n* points and *d* features, and $X_{test} \in \mathbb{R}^{m \times d}$ denote the test data with *m* points. The first step is to generate a pool of base detectors *C*, consisting of *r* detectors initialized with different parameters, e.g., a group of LOF detectors with distinct *n_neighbors* (also known as *MinPts* [7]). All base detectors are then trained and asked to predict on $X_{train}$, producing a train outlier score matrix $O(X_{train})$ shown as

Eq. (2), where $C_i(\cdot)$ denotes the score prediction function. Each base detector score $C_i(X_{train})$ is supposed to be normalized into a comparable scale, e.g. using Z-normalization [3, 38].

$$O(X_{train}) = [C_1(X_{train}),...,C_r(X_{train})] \in \mathbb{R}^{n \times r} \quad (2)$$

## 3.2 Model Selection and Combination

As DCSO needs to evaluate the detector competency when the ground truth is missing, two pseudo ground truth generation methods are introduced, in which the pseudo ground truth of $X_{train}$ is denoted as $target$: (i) averaging all base detector scores as shown in Eq. (3) and (ii) taking the maximum score across all detectors. Two DCSO methods are therefore designed: *DCSO_A* uses the pseudo training ground truth generated by averaging, while *DCSO_M* depends on the pseudo training ground truth by maximization. It should be noted that the pseudo ground truth here is for training data, which is therefore different from *SG_A* and *SG_M* that generate the scores for test instances instead.

$$target = \frac{1}{r}\sum_{i=1}^{r} C_i(X_{train}) \in \mathbb{R}^{n \times 1} \quad (3)$$

The local region of a test instance $X_{test\_i}$ is denoted as $\psi$, which consist of its $k$ nearest training objects by Euclidean distance. $k$NN is recommended for defining the local region over clustering since it has shown a better precision in DCS [11]. Euclidean distance $d_E$ is calculated using Eq. (4), where $p$ and $q$ are two equal-length vectors ($l$ is the vector length).

$$d_E(p,q) = \sqrt{\sum_{i=1}^{l}(p_i - q_i)^2} \quad (4)$$

Once the pseudo training ground truth $target$ and the local region $\psi$ are defined, the local pseudo target $target_k \in \mathbb{R}^{k \times 1}$ can be queried by selecting the points in $\psi$ from $target$. Similarly, the local training outlier score $O(X_{train\_k})$ can be easily acquired by selecting from the pre-calculated training score matrix $O(X_{train})$ as $O(X_{train\_k}) = [C_1(X_{train\_k}),...,C_r(X_{train\_k})] \in \mathbb{R}^{k \times r}$. Clearly, for different test objects, the local region needs re-calculating, but the local pseudo ground truth and the detector outlier scores can be queried from pre-calculated $target$ and $O(X_{train})$ efficiently.

For evaluating base estimator competency, DCS measures the local accuracy of base classifiers by the percentage of correctly classified points [20, 36], while DCSO measures the similarity between a base detector score to the pseudo target instead. This difference is caused by the lack of direct and reliable ways to gain binary labels in unsupervised outlier mining. Although converting pseudo outlier scores to binary labels is feasible, defining an accurate threshold for the conversion is challenging. Additionally, as outlier data is typically imbalanced, it is more stable to use similarity measures such as Pearson Correlation other than absolute accuracy or precision for competency evaluation. Therefore, DCSO measures the local competency of a base detector by the Pearson correlation between the local pseudo ground truth $target_k$ and the local detector score $C_i(X_{train\_k})$ as $\rho(target_k, C_i(X_{train\_k}))$ with Eq. (1), iterating over all $r$ base detectors. The detector with the highest Pearson Correlation $C_i^*$ is chosen as the most competent local detector for $X_{test\_i}$, and its prediction score $C_i^*(X_{test\_i})$ becomes the final score of $X_{test\_i}$.

## 3.3 Dynamic Outlier Ensemble Selection

Selecting only one detector, even if it is most similar to the pseudo ground truth, can be risky in unsupervised learning. However, this risk can be mitigated by selecting the top $s$ most similar detectors to the pseudo target for a second-phase combination instead of only using the most similar one. This idea can be viewed as an adaption of supervised DES [20] to outlier detection tasks. DCSO ensemble variations (*DCSO_MOA* and *DCSO_AOM*) are therefore introduced. Specifically, *DCSO_MOA* takes the maximum scores of top $s$ most similar detectors to the pseudo ground truth as the second-phase combination, while *DCSO_A* only take the most similar one. Similarly, *DCSO_AOM* additionally takes the average of $s$ selected detectors when the pseudo target is generated by maximization in *DCSO_M*.

---

**Algorithm 1** Dynamic Outlier Detector Combination (DCSO)

**Input**: the pool of detectors $C$, training data $X_{train}$, test data $X_{test}$, the local region size $k$

**Output**: outlier score for each test instance $X_{test\_i}$ in $X_{test}$

1. Train all base detectors in $C$ on $X_{train}$
2. Generate training outlier score matrix $O(X_{train})$ with Eq. (2)
3. **if** (*DCSO_A* or *DCSO_MOA*) **then**
4.    $target := avg(O(X_{train}))$ /* pseudo target by avg */
5. **else**
6.    $target := max(O(X_{train}))$ /* pseudo target by max */
7. **end if**
8. **for** each testing instance $X_{test\_i}$ in $X_{test}$ **do**
9.    Find its $k$ nearest neighbors in $X_{train}$ as $\psi$
10.    Get local pseudo target $target_k$ by selecting the subset
11.    of $k$ neighbors in $\psi$ from $target$
12.    **for** each base detector $C_i$ in $C$ **do**
13.      Let $C_i$ predict on $\psi$ to get outlier score $C_i(\psi)$
14.      Evaluate the local competency of $C_i$ by the similarity
15.      between $target_k$ and $C_i(\psi)$, e.g., using Eq. (1)
16.    **end for**
17.    **if** (*DCSO_A* or *DCSO_M*) **then** /*select the best one*/
18.      Select the most similar detector $C_i^*$
19.      Use $C_i^*(X_{test\_i})$ as the output of test instance $X_{test\_i}$
20.    **else** /* DCSO ensemble for second-phase combination */
21.      Select $s$ most similar detectors dynamically and add
22.      to set $C_s^*$, e.g., using Eq. (5) - (6)
23.      **if** (*DCSO_AOM*) **then**
24.        Take the average of $C_i^*(X_{test\_i})$ for detectors
25.        in $C_s^*$ as the final score of $X_{test\_i}$
26.      **else** /* *DCSO_MOA* */
27.        Take the maximization of $C_i^*(X_{test\_i})$ for detectors
28.        in $C_s^*$ as the final score of $X_{test\_i}$
29.      **end if**
30.    **end if**
31. **end for**

## 3.4 The Similarities and Differences between SG and DCSO Methods

The workflow of all four DCSO methods is shown in Algorithm 1 and the critical differences between static global algorithms and dynamic combination algorithms are demonstrated in Figure 1. It is apparent that the correlation among detectors on $k$ training points is different from the global correlation based on all $n$ training points. This differentiates DCSO from simple global averaging, as the former stresses the importance of the local data relationship. Besides, global averaging considers all base detectors with either equal weights (*SG_A*) or different weights (*SG_WA*), while DCSO is "winner-takes-all"—only the best detector is kept and all the rest are discarded. In terms of *DCSO_AOM* and *DCSO_MOA*, they can be viewed as dynamic local variations of *SG_AOM* and *SG_MOA*. However, the difference is subtler than merely local versus global algorithms; it also lies in how the constituent detectors are selected. *SG_AOM* and *SG_MOA* build subgroups by selecting base detectors randomly, while *DCSO_AOM* and *DCSO_MOA* select competent detectors by similarity rank with less uncertainty. Compared with *SG_AOM* and *SG_MOA*, DCSO ensembles can show how the prediction is made for each test object, which enhances the model reliability. Table 1 presents some intuitive connections between the selected SG algorithms and our proposed DCSO algorithms.

## 3.5 The Impact of Parameters and Competency Evaluation Functions

In the **Combination** stage, $k$ decides the number of nearest neighbors to consider while defining the local region. Small $k$ implies more attention to the local relationship, while large $k$ makes it more global. When $k$ equals $n$, the number of training points, DCSO is similar to static global methods; $k$ therefore should not be too large. Additionally, large $k$ leads to higher computational cost. Nonetheless, the local region size should not be too small, because small $k$ can be problematic for Pearson correlation calculation due to low stability. Unlike supervised learning that can possibly determine optimal $k$ by cross validation [20], unsupervised learning does not have a trivial way to decide.

For ensemble variation DCSO (*DCSO_MOA* and *DCSO_AOM*), the number of selected base detectors, $s$, affects the strength of dynamic. Setting $s=1$ produces exactly the original DCSO algorithms (*DCSO_A* and *DCSO_M*); increasing $s$ from 1 to $r$, the total number of base detectors, results in more static algorithms. A varying $s$ is recommended over a fixed $s$ for better flexibility. Let $\phi(X_{test\_i})$ in Eq. (5) denote the Pearson correlation of all base detectors for $X_{test\_i}$, only the detectors exceeding the correlation threshold $\theta$ in Eq. (6) are selected for second-phase combination. Equation (6) calculates the local correlation threshold by finding the values within $\alpha$ times standard deviation from the highest correlation. Larger selection strength factor $\alpha$ leads to larger $s$; it indirectly controls the strength of dynamic in a more flexible way.

$$\phi(X_{test\_i}) = [\rho(C_1(X_{train\_k}, target_k), ..., \rho(C_d(X_{train\_k}, target_k)] \quad (5)$$

$$\theta = \phi(X_{test\_i})_{\max} - \alpha \cdot \phi(X_{test\_i})_\sigma \quad (6)$$

As for local competency evaluation, there are alternative methods in addition to Pearson correlation, such as widely used Euclidean distance shown in Eq. (4). Moreover, both Pearson correlation and Euclidean distance could be used along with weights, such as weighted Pearson correlation. Rayana and Akoglu used the outlyingness rank of the pseudo target as weights [30]. Alternatively, Ko et al. used the Euclidean distance between the test instance and its $k$ nearest neighbors as weights [20]. The former gives more attention to outliers whereas the latter stresses the importance of locality.

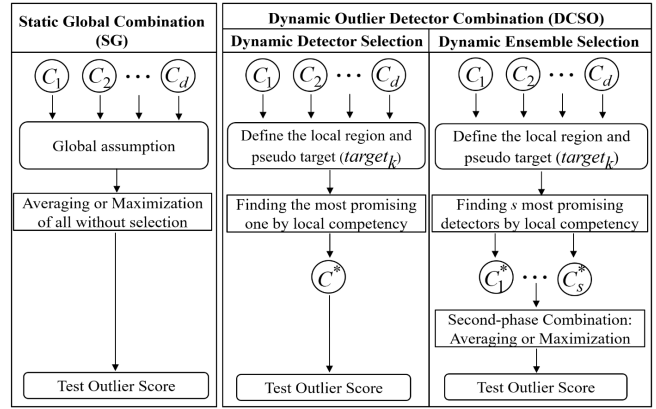

**Figure 1. Workflows of SG and DCSO methods**

Kd-tree can speed up the local region definition by $k$NN for low dimensional spaces, and prototype selection and fast approximate methods can expedite $k$NN for more complicated feature spaces [11, 18]. With the appropriate implementation of *DCSO_A* and *DCSO_M*, the time complexity for each test instance is $O(nd + n\log(n))$; $O(nd)$ is for the distance calculation, and $O(n\log(n))$ is for summation and sorting [20]. To combine $s$ base detectors in *DCSO_MOA* and *DCSO_AOM*, additional $O(s)$ is needed, resulting in $O(nd + n\log(n) + s)$ time complexity.

## 3.6 Theoretical Analysis

It has been shown that combining diversified base detectors, such as averaging, results in variance reduction [3, 30, 31]. However, simply combining all base detectors may also include inaccurate ones, leading to higher bias. This explains why static global averaging does not work well due to high bias. In contrast, reducing bias in unsupervised outlier ensemble is not trivial [2, 31]. With Aggarwal's bias-variance framework, DCSO can be regarded as a combination of both variance and bias reduction. DCSO initializes various base detector with different parameters to induce diversity. *DCSO_A* use the pseudo target generated by averaging that leads to an indirect variance reduction. DCSO focuses on the local competency evaluation, which helps to find the base detectors with low model bias. Additionally, *DCSO_M* is more stable than global maximization (*SG_M*), as the variance is reduced by using the most competent detector's output other than using global maximum values of base detectors. In addition to the benefits of *DCSO_A* and *DCSO_M*, *DCSO_MOA* and *DCSO_AOM* have a second-phase combination (maximization or averaging) to further decrease the generalization error through bias reduction and variance reduction, respectively. Thus, DCSO may reduce the generalization error through both variance and bias reduction channels. Despite, DCSO is a heuristic framework, and the result could be unpredictable on pathological datasets.

**Table 1. The connections between SG and DCSO methods**

| Combination or Pseudo Target Generation | Static Global Methods | Dynamic Methods |
|---|---|---|
| Averaging | SG_A | DCSO_A |
| Maximization | SG_M | DCSO_M |
| Maximum-of-Average | SG_MOA | DCSO_MOA |
| Average-of-Maximization | SG_AOM | DCSO_AOM |

## 4. NUMERICAL EXPERIMENTS
## 4.1 Datasets and Evaluation Metrics
Table 2 shows ten outlier datasets used in this study that are openly accessible at [27]. All datasets are randomly split at 60% for training and 40% for testing. The average scores of 20 independent trials are used for evaluation. Multiple comparison analyses are conducted, in which the area under the receiver operating characteristic (ROC) curve and precision at rank $m$ ($P_{@m}$) are used for evaluation. Both metrics are widely used in outlier research [2, 4, 25, 30, 37]. Non-parametric Wilcoxon rank-sum test [34] is used to determine whether two results have a significant difference. For multi-group comparison, non-parametric Friedman test [16] followed by post-hoc test, Nemenyi test [26], is used. For all these tests, $p < 0.05$ is considered to be statistically significant, otherwise non-significant.

## 4.2 Base Detector Initialization
To test the applicability of DCSO, two unsupervised outlier detection methods, LOF and $k$NN, are used to construct the pool of base detectors, respectively. For a test instance, $k$NN regards the Euclidean distance between the instance and its $k$th nearest training point as the outlier score. Clearly, the $k$NN detector herein has a distinct usage from the $k$NN used for defining the local region. To induce diversity among base detectors, distinct initialization parameters are used. For both LOF and $k$NN, the number of neighbors, $n\_neighbors$, varies in the range of $[10, 20, ..., 200]$, resulting in 20 diversified base detectors.

## 4.3 Experiment Design
Experiment I compares six SG algorithms introduced in Section 2.3 with four proposed DCSO algorithms shown in Table 1 and Algorithm 1. For *SG_AOM* and *SG_MOA*, 5 subgroups are built, and each subgroup contains 4 base detectors without replacement. For all DCSO algorithms, $k$, the local region size, is fixed at 100 for consistency; selection strength factor $\alpha$ in Eq. (6) is set to 0.2 for DCSO ensemble methods (*DCSO_MOA* and *DCSO_AOM*).

In this study, Z-normalization is applied first to eliminate the scale difference among various base detector scores before the combination [3, 38]. Z-normalization shown in Eq. (7) can scale a vector $\boldsymbol{x}$ to zero mean ($\mu = 0$) and unit variance ($\sigma = 1$).

$$Z(x_i) = \frac{x_i - \mu}{\sigma} \quad (7)$$

Experiment II compares the performances between Pearson correlation and Euclidean distance for local detector competency evaluation. The effects of three choices of weight are analyzed as well: (i) outlyingness rank [30]; (ii) Euclidean distance [20] and (iii) none. It is noteworthy that smaller Euclidean distance between two data points implies higher weight, while smaller Pearson correlation implies lower weight. For the sake of consistency, the weight by Euclidean distance is inverted using $w_i = max(\boldsymbol{d}) - d_i$, where $\boldsymbol{d}$ denotes a Euclidean distance vector.

## 5. RESULTS AND DISCUSSIONS
Due to its great compatibility, DCSO can work with both LOF and $k$NN base detectors. Table 3 and 4 show the results of ROC and $P_{@m}$ on ten datasets, in which LOF is used as the base detector. The highest score is highlighted in bold, while the lowest is marked with an asterisk. The experimental results when $k$NN is used as the base detector are accessible online for brevity[1]. The analyses of both $k$NN and LOF detector demonstrate that DCSO can bring consistent performance improvement over its SG counterparts, which is especially significant regarding $P_{@m}$.

[1] https://github.com/yzhao062/DCSO

**Table 2. Real-world datasets used for evaluation**

| Dataset [27] | Pts ($n$) | Dim ($d$) | Outlier | % Outlier |
|---|---|---|---|---|
| Pima | 768 | 8 | 268 | 34.89 |
| Vowels | 1456 | 12 | 50 | 3.434 |
| Letter | 1600 | 32 | 100 | 6.250 |
| Cardio | 1831 | 21 | 176 | 9.612 |
| Thyroid | 3772 | 6 | 93 | 2.466 |
| Satellite | 6435 | 36 | 2036 | 31.64 |
| Pendigits | 6870 | 16 | 156 | 2.271 |
| Annthyroid | 7200 | 6 | 534 | 7.417 |
| Mnist | 7603 | 100 | 700 | 9.207 |
| Shuttle | 49097 | 9 | 3511 | 7.151 |

## 5.1 Algorithm Performances
The Friedman test shows there is a statistically significant difference of ten algorithms regarding $P_{@m}$ ($\chi^2 = 29.71$, $p = 0.0005$). Despite, the Nemenyi test fails to spot specific pairs of algorithms with a significant difference due to its weak power [13], which is acceptable given the limited number of datasets. In general, DCSO algorithms show great potential: they achieve the highest ROC score on eight out of ten datasets, and the highest $P_{@m}$ score on all datasets. The improved $P_{@m}$ is possibly due to DCSO's strong ability to find local outliers, even at the expense of misclassifying few normal points, leading to slightly lower ROC occasionally. Specifically, *DODS_MOA* is the most performing method that ranks highest on five datasets and second highest on two datasets for both ROC and $P_{@m}$. As for static global methods, *SG_M* achieves the highest ROC on **Vowels** and **Letter** but ranks the lowest on **Thyroid**. Other static global algorithms never achieve the highest score on any dataset besides *SG_AOM* shows the highest $P_{@m}$ on **Annthyroid**. It should be noted that the SG methods with a second-phase combination (*SG_MOA* and *SG_AOM*) show better performance than *SG_A*, and better stability than *SG_M*. The observations of SG methods well agree with the conclusions in Aggarwal's work [2, 3]: (i) averaging (*SG_A*) could reduce variance but may lose the high performing ones, and get closer to random outlier score with the increasing dimension; (ii) maximization (*SG_M*) has bias reduction effect and is good at identifying "well hidden" outliers at the expense of increased variance, leading to unstable results and (iii) *SG_AOM* and *SG_MOA* outperfom since they leverage both bias and variance reduction through the second-phase combination.

*DCSO_A* and *DCSO_M* do not show superiority to their static global counterparts. It is understood that the pseudo ground truth is unlikely to be accurate with inherent bias, causing inaccurate local competency evaluation. *DCSO_A* uses the pseudo ground truth generated by averaging all detector scores. It achieves the highest score on **Pima** and **Thyroid**, whereas ranks the lowest on four datasets. Theoretically, using averaged scores as the pseudo ground truth indirectly benefits from the variance reduction effect, and concentrating on the local region to select the most competent detector reduces the model bias. However, *DCSO_A* only selects the most competent one and discards all other detectors, yielding a weaker variance reduction effect than *SG_A* that uses all detector scores. The weak variance reduction may not offset the inherent bias of the pseudo ground truth, leading to poor results. As discussed, *SG_M* has unstable performances that vary drastically on different datasets. As for *DCSO_M*, it uses the maximization score as the pseudo ground truth and exhibits the unstable behavior like *SG_M*; it is even inferior to *SG_M* at times. Clearly, the pseudo ground truth generated by maximization also comes with high variance, and the model variance is further increased since *DCSO_M* focuses on the local region. For static global methods, Aggarwal has shown that averaging leads to limited performance improvement, while maximization is riskier but with potentially higher gains [3]. However, the difference is less significant when they are used as the pseudo ground truth for *DCSO_A* and *DCSO_M*; a Wilcoxon rank-sum test shows no significance between two generation methods. One explanation is that DCSO only uses the pseudo ground truth to find the most similar detector, and all unpicked detectors are discarded. Thus, the characteristics of the selected detector are kept instead of being neutralized in *SG_A* or polarized in *SG_M*. This self-adaptive mechanism downplays the importance of pseudo ground truth generation methods. In addition, both generation methods are heuristic with unpredictable accuracy—it is not surprising to observe close performances between *DCSO_A* and *DCSO_M*.

Table 3. ROC performances (average of 20 independent trials, highest score highlighted in bold, lowest score marked with *)

| Dataset | SG_A | SG_M | SG_WA | SG_THRESH | SG_AOM | SG_MOA | DCSO_A | DCSO_M | DCSO_MOA | DCSO_AOM |
|---|---|---|---|---|---|---|---|---|---|---|
| Pima | 0.6897 | 0.6542 | 0.6907 | 0.6285* | 0.6777 | 0.6836 | **0.6957** | 0.636 | 0.6911 | 0.6375 |
| Vowels | 0.9116 | **0.9302** | 0.9096* | 0.9229 | 0.9213 | 0.9178 | 0.9209 | 0.9237 | 0.9146 | 0.9248 |
| Letter | 0.7783 | **0.8481** | 0.7737 | 0.7980 | 0.8117 | 0.8016 | 0.7553* | 0.8469 | 0.7866 | 0.8456 |
| Cardio | 0.9062 | 0.8939 | 0.9077 | 0.9087 | 0.9169 | 0.9166 | 0.9017 | 0.8973 | **0.9179** | 0.8783* |
| Thyroid | 0.9691 | 0.9389* | 0.9700 | 0.9679 | 0.9616 | 0.9657 | **0.9712** | 0.9427 | 0.9594 | 0.9438 |
| Satellite | 0.6001 | 0.6391 | 0.5995 | 0.6204 | 0.6313 | 0.6204 | 0.5949* | 0.6180 | **0.6412** | 0.6163 |
| Pendigits | 0.8399 | 0.8587 | 0.8443 | 0.8564 | 0.8694 | 0.8574 | 0.8416 | 0.8701 | **0.8867** | 0.8035* |
| Annthyroid | 0.7684 | 0.7869 | 0.7657 | 0.7639 | 0.7765 | 0.7752 | 0.7545* | 0.7967 | 0.7561 | **0.7987** |
| Mnist | 0.8518 | 0.8417 | 0.8525 | 0.8243 | 0.8606 | 0.8580 | 0.8532 | 0.8138 | **0.8658** | 0.8056* |
| Shuttle | 0.5388 | 0.5534 | 0.5388 | 0.5448 | 0.5514 | 0.5441 | 0.5327* | 0.5329 | **0.5682** | 0.5341 |

Table 4. $P_{@m}$ performances (average of 20 independent trials, highest score highlighted in bold, lowest score marked with *)

| Dataset | SG_A | SG_M | SG_WA | SG_THRESH | SG_AOM | SG_MOA | DCSO_A | DCSO_M | DCSO_MOA | DCSO_AOM |
|---|---|---|---|---|---|---|---|---|---|---|
| Pima | 0.5100 | 0.4683 | 0.5127 | 0.4933 | 0.4957 | 0.5039 | **0.5175** | 0.4576 | 0.5083 | 0.4576* |
| Vowels | 0.3074 | 0.3250 | 0.3029* | 0.3074 | 0.3302 | 0.3185 | **0.3682** | 0.3044 | 0.3395 | 0.3161 |
| Letter | 0.2508 | 0.3547 | 0.2469 | 0.2508 | 0.2950 | 0.2699 | 0.2426* | **0.3795** | 0.2862 | 0.3785 |
| Cardio | 0.3601 | 0.3733 | 0.3624 | 0.3728 | 0.4233 | 0.4104 | 0.3553 | 0.3676 | **0.4453** | 0.3201* |
| Thyroid | 0.3936 | 0.2589 | 0.4061 | 0.3968 | 0.3731 | 0.3896 | **0.4182** | 0.2080* | 0.3730 | 0.2449 |
| Satellite | 0.4301* | 0.4500 | 0.4306 | 0.4466 | 0.4480 | 0.4414 | 0.4400 | 0.4427 | **0.4509** | 0.4398 |
| Pendigits | 0.0733 | 0.0590 | 0.0709 | 0.0700 | 0.0637 | 0.0617 | 0.0749 | 0.0595 | **0.0811** | 0.0560* |
| Annthyroid | 0.2943 | 0.2951 | 0.2975 | 0.2997 | **0.3215** | 0.3103 | 0.3065 | 0.2904* | 0.3075 | 0.3046 |
| Mnist | 0.3936 | 0.3737 | 0.3944 | 0.3956 | 0.3966 | 0.3976 | 0.3973 | 0.3541 | **0.4123** | 0.3520* |
| Shuttle | 0.1508 | 0.1484 | 0.1434 | 0.1582 | 0.1591 | 0.1600 | 0.1589 | 0.1389* | **0.1604** | 0.1393 |

Dynamic ensemble selection with a second-phase combination (*DCSO_MOA* and *DCSO_AOM*) may overcome the limitations of *DCSO_A* and *DCSO_M*. A Friedman test reveals there is a significant difference among four DCSO algorithms regarding $P_{@m}$ ($\chi^2 = 13.24$, $p = 0.004$). As discussed in [2, 3], averaging outlier scores often loses both highly performing and poorly performing ones, which produces mediocre results. Conducting a second-phase maximization could therefore mitigate this risk, leading to a low bias model. *DCSO_MOA* takes the maximization of the selected detectors, which could be viewed as a further reduction of the model bias over *DCSO_A*. This helps when the pseudo ground truth has limited accuracy. The results show *DCSO_MOA* have better ROC and $P_{@m}$ on eight out of ten datasets than *DCSO_A*, and the $P_{@m}$ improvement is especially significant on **Letter** (17.97%) and **Cardio** (25.33%). *DCSO_MOA* also outperforms its static global counterpart (*SG_MOA*) regarding $P_{@m}$ on eight out of ten datasets, especially significant on **Pendigits** (31.44%). In contrast, the benefit of taking the second-phase combination is less effective for *DCSO_AOM*, which is not superior to neither *DCSO_M* nor *SG_AOM*. Theoretically, *DCSO_AOM*'s concentration on the local competency evaluation could improve the model bias, and the second-phase averaging would decrease the model variance leading to a stability improvement effect. However, the variance reduction effect of the additional averaging does not offset the bias increase by local competency evaluation in *DCSO_AOM* and its inherent high instability of the pseudo ground truth generated by maximization. Thus, only *DCSO_MOA* is suggested for detector combination among all DCSO algorithms, leading to more stable and superior results due to *DCSO_MOA*'s effective combined bias and variance reduction capacity.

## 5.2 Visualization Analysis

Figure 2 visually compares the performance of the most competent SG and DCSO methods on **Cardio**, **Thyroid** and **Letter** using t-distributed stochastic neighbor embedding (t-SNE) [24]. The green and blue markers highlight objects that can only be correctly classified by either the SG or DCSO methods, respectively, to emphasize the mutual exclusivity of the two approaches. The visualizations of **Cardio** (left) and **Thyroid** (middle) illustrate that DCSO methods have an edge over SG methods in detecting local outliers when they cluster together (highlighted by red dotted circles in Fig. 2). Additionally, DCSO methods can contribute to classifying both outlying and normal points as long as the data locality matters. However, outlying data distribution on **Letter** (right) is more dispersed—outliers do not form local clusters but mix with the normal points. This causes DCSO is slightly inferior to *SG_M* regarding ROC, although DCSO still shows a $P_{@m}$ improvement. Based on the visualizations, some assumptions can be made. Firstly, DCSO is useful when outlying and normal objects are well separate, but less effective when they are interleaved with increased difficulty to form local clusters. Secondly, the local region size $k$ defines the data locality and therefore impacts the effectiveness of DCSO. A Friedman test shows a significant different regarding different $k$ choices (10, 30, 60 and 100) for both ROC ($\chi^2 = 13.27$, $p = 0.004$) and $P_{@m}$ ($\chi^2 = 20.98$, $p = 0.0001$). For instance, the total number of outliers in the testing set (40% of the entire dataset) of **Vowels** and **Letter** is only 20 and 40, respectively, which may not be sufficient to form local outlier clusters since $k$ is set to 100 in this study. A smaller $k$ is more appropriate when a limited number of outliers is assumed. Thirdly, defining the data locality using $k$NN can be problematic in high-dimensional space since many irrelevant features may be presented [4]. The datasets with a relatively large number of features have a high possibility of including irrelevant features. This may explain why DCSO is less performing on **Letter** ($d = 32$) and **Mnist** ($d = 100$).

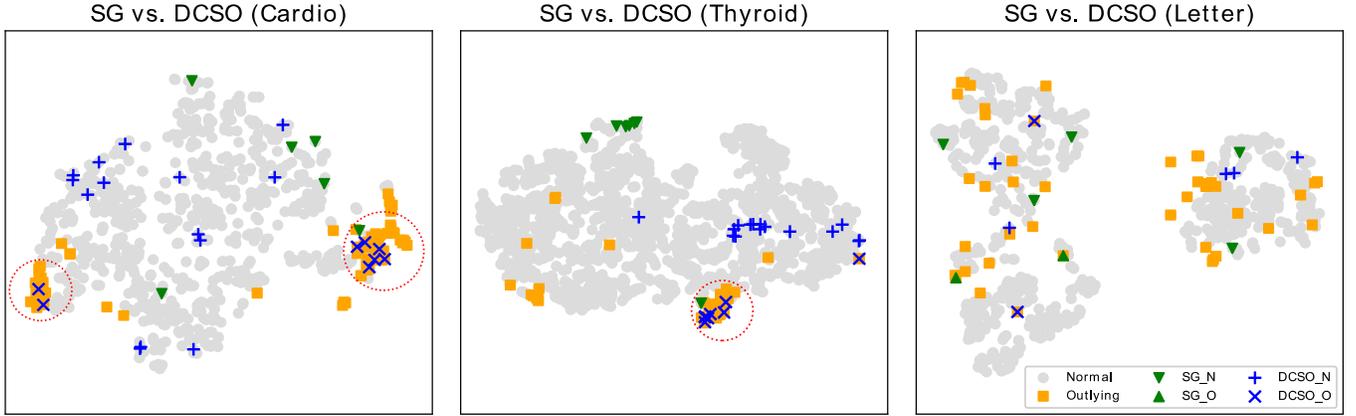

**Figure 2. t-SNE visualizations on Cardio (left), Thyroid (middle) and Letter (right), where normal and outlying points are denoted as grey dots and orange squares, respectively. The normal and outlying points that can only be correctly identified by SG methods are labeled as *SG_N* (green triangle-down) and *SG_O* (green triangle-up). Similarly, the normal and outlying points that can only be correctly identified by DCSO methods are labeled as *DCSO_N* (blue plus sign) and *DCSO_O* (blue cross sign).**

## 5.3 Competency Evaluation Methods

Local competency evaluation depends on the similarity measure among the pseudo ground truth and base detector scores. It is noted that a Friedman test carried out to compare Pearson correlation and Euclidean distance does not reveal significant difference regarding ROC and $P_{@m}$. Numerical analysis shows that the performance difference is often negligible (within 1%). In addition, a separate Friedman test shows that the choice of weight (outlyingness rank, Euclidean distance and none) does not pose an impact on model performance. Thus, it is unnecessary to use weighted similarity measure that has higher computational cost.

The observations are understandable because Pearson correlation is equivalent to Euclidean distance while measuring the similarity between two normalized vectors [5]. Equation (8) shows the equivalence for two equal-length normalized vectors $p$ and $q$ ($l$ denotes vector length). DCSO applies Z-normalization to each detector scores first, so $C_i(X_{train})$ is normalized regarding all $n$ training points. In this study, the local region size $k$ is set to a relatively large value 100, so $target_k$ and $C_i(X_{train\_k})$ can be considered approximately normalized, unless the local region contains a lot of outliers. This explains why two similarity measures have close performances. Despite, when a relatively small $k$ is chosen, Eq. (8) may fail to apply, since the local pseudo target and the local detector scores on $k$ instances are less likely to be normalized. Therefore, when $k$ value is large, the most efficient method, Euclidean distance without considering the weight, is recommended as the similarity measure among the pseudo ground truth and base detectors scores for lower computational cost.

$$d_E(p, q) = \sqrt{2l \cdot \rho(p, q)} \qquad (8)$$

## 5.4 Limitations and Future Directions

Numerous investigations are underway. Firstly, the local region is defined as the $k$ nearest training data points of the test instance. However, it is not ideal due to: (i) high time complexity [11]; (ii) the lack of a reliable $k$ setting criteria and (iii) degraded performance when many irrelevant features are presented in high dimensional space [4]. This may be improved by using prototype selection [11], fast approximate methods [18] or even defining the local region by advanced clustering methods instead [11]. Secondly, only simple pseudo ground truth generation methods are explored (averaging or maximization) in this study; more complicated and accurate methods should be considered, such as actively pruning base detectors [30]. Lastly, DCS has proven to work with heterogeneous base classifiers in classification problems [11, 20], which is pending for verification in DCSO. More significant improvement is expected, as the base detectors used in this study are homogeneous with limited diversity.

## 6. CONCLUSIONS

A new and improved unsupervised framework called DCSO (Dynamic Combination of Detector Scores for Outlier Ensembles) is proposed and assessed in the selection and combination of base outlier detector scores. Unlike traditional ensemble methods that combine constituent detectors statically, DCSO dynamically identifies top-performing base detectors for each test instance by evaluating detector competency in its defined local region. Given the fact that local relationships for data are critical in outlier score combination, DCSO ranks the competency of individual base detectors by its similarity to the pseudo ground truth in the local region. To improve model stability and reduce the risk of using an individual detector, ensemble variations of DCSO are also provided in this research.

The proposed DCSO framework is assessed using statistical evaluation techniques on ten real-world datasets. The results of this evaluation validate the effectiveness of the DCSO framework in detecting outliers over traditional static combination methods. In addition to markedly improved outlier detection capabilities, DCSO is also computationally robust in that is it compatible with any base detectors (e.g. LOF or $k$NN) and transparent in showing how outlierness scores are generated for each test instance by providing the selected base detector. Theoretical considerations are also provided for DCSO, alongside complexity analyses and visualizations, to provide a holistic view of this unsupervised outlier ensemble method. Moreover, the effect of parameter selection is discussed and empirical parameter setting instructions are provided. Lastly, all source codes, experiment results and figures used in this study are made publicly available [1].

---

[1] https://github.com/yzhao062/DCSO